\UseRawInputEncoding
\documentclass[conference]{IEEEtran}
\IEEEoverridecommandlockouts
\usepackage{cite}
\usepackage{amsmath,amssymb,amsfonts}
\usepackage{algorithmic}
\usepackage{graphicx}
\usepackage{textcomp}
\usepackage{xcolor}
\def\BibTeX{{\rm B\kern-.05em{\sc i\kern-.025em b}\kern-.08em
    T\kern-.1667em\lower.7ex\hbox{E}\kern-.125emX}}
\begin{document}

\title{End-to-end Deep Object Tracking\\ with Circular Loss Function\\ for Rotated Bounding Box\\
}
\author{
\IEEEauthorblockN{Vladislav Belyaev}
\IEEEauthorblockA{JetBrains Research\\
National Research University\\
Higher School of Economics\\
St. Petersburg, Russia \\
belyaev.vladislav.nw@gmail.com}
\and
\IEEEauthorblockN{Aleksandra Malysheva}
\IEEEauthorblockA{JetBrains Research\\
National Research University\\
Higher School of Economics\\
St. Petersburg, Russia \\
malyshevasasha777@gmail.com}
\and
\IEEEauthorblockN{Aleksei Shpilman}
\IEEEauthorblockA{JetBrains Research\\
National Research University\\
Higher School of Economics\\
St. Petersburg, Russia \\
alexey@shpilman.com}
}

\maketitle

\begin{abstract}
The task object tracking is vital in numerous applications such as autonomous driving, intelligent surveillance, robotics, etc. This task entails the assigning of a bounding box to an object in a video stream, given only the bounding box for that object on the first frame. In 2015, a new type of video object tracking (VOT) dataset was created that introduced rotated bounding boxes as an extension of axis-aligned ones. In this work, we introduce a novel end-to-end deep learning method based on the Transformer Multi-Head Attention architecture. We also present a new type of loss function, which takes into account the bounding box overlap and orientation.

Our Deep Object Tracking model with Circular Loss Function (DOTCL) shows an considerable improvement in terms of robustness over current state-of-the-art end-to-end deep learning models. It also outperforms state-of-the-art object tracking methods on VOT2018 dataset in terms of expected average overlap (EAO) metric.
\end{abstract}

\begin{IEEEkeywords}
visual tracking, transformer, siamese networks
\end{IEEEkeywords}

\section{Introduction}
Video object tracking is essential to many areas of research and industry including but not limited to autonomous driving, intelligent surveillance, and robotics.

The standard formulation for the object tracking is assigning a so called bounding box to an object in a video stream (i.e., on every frame of the video), provided only the bounding box for a chosen object on the first frame. VOT challenges and corresponding datasets \cite{kristan2018sixth} are a generally accepted benchmark for this task. These datasets consist of several video files with manually assigned bounding boxes for one selected object. Videos vary significantly in both the filming conditions (i.e., the scene lighting) and the object size relative to the frame. 

In 2015, organizers of the VOT challenge compiled a new type of video object tracking dataset that introduced rotated bounding boxes as opposed to axis-aligned bounding boxes. This was motivated by the fact that axis-aligned bounding boxes approximated the target with percentage of pixels within the bounding box at average of $45\%$ but the rotated box can increase this percentage to $60\%$ \cite{kristannovel}. But the introduction of the rotated bounding boxes complicates the task even further. 

Most state-of-the art trackers fall into two main categories. 

The first group of methods applies correlation filters (CF) to cyclic matrices to estimate the probability of the object in sliding windows in the vicinity of the previous location \cite{wallace1980analysis}. CF based trackers may calculate correlation between various features such as pixel values, HOG features~\cite{lukezic2017discriminative}, and even values from different layers of a deep convolutional neural network (CNN~\cite{bai2018multi}). The CNN for feature extraction is typically pretrained on a object classification problem on a large dataset, such as ImageNet~\cite{russakovsky2015imagenet}.

The second group trains an end-to-end deep neural network model to output the bounding box directly. Parts of these networks can also be pretrained on a large dataset for the classification task such as ImageNet~\cite{russakovsky2015imagenet}.

Most of the top trackers of the VOT2018 competition \cite{kristan2018sixth} use only the last and the current frame for tracking and do not attempt to extract any additional information from the sequence of frames. Some recent approaches, such as Recurrent YOLO (ROLO) \cite{ning2017spatially} and Deep Reinforcement Learning for Visual Object tracker (DRLT) \cite{zhang2017deep}, use historical visual semantics but do not demonstrate competitive performance.

In this work,  we propose a novel end-to-end deep learning visual object tracking method. It applies the Transformer Multi-Head Attention approach~\cite{vaswani2017attention} as well as a new type of loss function. Transformer architecture efficiently analyses whole sequences of frames and bounding boxes. The new loss function takes into account the bounding box overlap and orientation and provides calculatable derivative for training the network with backpropagation algorithm.

\section{Related work}

As mentioned above, we can divide the existing state-of-the-art trackers into two main groups. The first group includes variations of the discriminative correlation filter applied to circular matrices. The second type is end-to-end deep learning trackers that take a frame or a series of frames as an input and directly output the bounding box. 

In this section, we describe current state-of-the-art methods in both groups that we use to compare with the performance of our algorithm. We present the top three performing models from the VOT2018 competition as baselines. Two of these models are discriminative correlation filters, and one is an end-to-end deep learning model.

\subsection{Discriminative Correlation Filters}\label{sec:dcf}

Discriminative Correlation Filter trackers (DCF~\cite{lukezic2017discriminative}) consider cyclic shifts of the original sample and calculate the correlation for maps of features in sliding windows around the object's previous location. Top two results in VOT2018 \cite{kristan2018sixth} competition are  variants of DCF that use both manually defined features (i.e., HOG and Color Names ~\cite{lukezic2017discriminative}) and features generated by CNN. Here we only give a brief characterization of both approaches, since our main contribution is in the area of end-to-end deep learning trackers. For more detailed descriptions, we refer the reader to corresponding papers \cite{bai2018multi, xu2018learning}. 

\subsubsection{Multi-hierarchical Independent Correlation Filter}

Multi-hierarchical Independent Correlation Filter (MFT) \cite{bai2018multi} showed the best performance in terms of the robustness (R) metric and second to best performance in terms of expected average overlap (EAO) metric in VOT2018 challenge. MFT tracker learns a hierarchy of multi-resolution deep features from the convolutional neural network (CNN) for correlation filters. These multi-hierarchical deep features of CNN representing different semantic information assist with tracking of the multi-scale objects. To overcome the deep feature redundancy, each hierarchical feature is independently fed into a single branch of correlation filters optimization to implement the online learning of parameters of these filters. Finally, an adaptive weighting scheme is integrated into the framework to fuse these independent multi-branch correlation filters to increase robustness.

\subsubsection{Learning Adaptive Discriminative Correlation Filter}

The top performer in the VOT2018 challenge by the primary metric, expected average overlap (EAO) \cite{kristan2015visual}, is the Learning Adaptive Discriminative Correlation Filter (LADCF) \cite{xu2018learning}. LADCF addresses the problem of spatial boundary effect and temporal filter degeneration. This model combines consistent temporal constraints and adaptive spatial regularization, which enables joint spatiotemporal filter learning. Also, the optimization framework is proposed to learn discriminative filters with the augmented Lagrangian method. 

\subsection{End-to-end Deep Learning Models}\label{sec:end-to-end}

The second group of object trackers is the end-to-end deep learning models. These models do not perform additional operations over the learned features of the pretrained neural network, but rather train the network to output the bounding box parameters directly. Siamese Region Proposal Network (SiamRPN) represents state-of-the-art methods of this group.

\subsubsection{Siamese Region Proposal Network}

The Siamese Region Proposal Network (SiamRPN) \cite{li2018high} consists of a classification branch and a regression branch. It makes use of a region proposal network first introduced in~\cite{ren2015faster} to propose several potential regions for an object. Classification branch is in charge of the object-background classification. Regression branch refines the bounding box proposal.

Authors train this network in the end-to-end fashion with the following loss function:

\begin{equation}
    \label{eq:loss_total}
    L_{\text{total}} = L_{\text{cls}} + \lambda L_{\text{reg}}
\end{equation}

where $\lambda$ is a hyper-parameter to balance the two loss functions parts. $L_{\text{cls}}$ is the cross entropy classification loss and $L_{\text{reg}}$ is the regression loss.

They define the regression loss $L_{\text{reg}}$ in the following manner:

\begin{equation}
    \label{eq:loss_reg}
    L_{\text{reg}} = \sum_{i = 0}^{3} \text{Smooth}_{L_1}(\delta[i], \sigma),
\end{equation}

where $\delta[i]$ is defined as:

\begin{align}
\delta[0] &= \frac{x_\text{gt} - x}{w},  &  \delta[1] &= \frac{y_\text{gt} - y}{h}, \\
\delta[2] &= \ln{\frac{w_\text{gt}}{w}},  &  \delta[3] &= \ln{\frac{h_\text{gt}}{h}},
\end{align}

where $x, y, w, h$ are the center point coordinates and the width and the height of the proposed bounding box and $x_\text{gt}, y_\text{gt}, w_\text{gt}, h_\text{gt}$ as those of the ground truth bounding box.

$\text{Smooth}_{L_1}$ is defined as:

\begin{equation}
\label{eq:smooth}
\text{Smooth}_{L_1}(z, \sigma) = \left\{ \begin{array}{ll}
 0.5 \sigma^2 z^2  & \text{if} |z| > \frac{1}{\sigma^2}\\
 |z| - \frac{1}{2 \sigma^2} & \text{otherwise}\\
  \end{array} \right.
\end{equation}

\subsubsection{RNN-based Object Trackers}

Most state-of-the-art models use object location history obtained from just one previous frame to search for an object. 

One of the intuitions behind this work is that historical visual semantics may contain valuable information for effective frame handling and continuous tracking. Thus, it is necessary to develop architectures that process sequences of frames rather than a single frame. Solutions such as Recurrent YOLO (ROLO) \cite{ning2017spatially} and Deep Reinforcement Learning for Visual Object tracker (DRLT) \cite{zhang2017deep} use Long-Term Short Memory modules (LSTM) \cite{hochreiter1997long} to analyze the sequence of frames to extract spatiotemporal information that is important for better tracking. 
However, as we previously mentioned, RNN trackers do not yet demonstrate competitive results.

\subsection{Transformer and Multi-head Attention}\label{sec:transformer}

The relatively new Transformer architecture \cite{vaswani2017attention} has demonstrated outstanding results in many natural language processing tasks and it proved itself to be well suited for the extraction of the complex dependencies between sequential elements. In its core, it is based on the attention mechanism, that have become one of the most prolific methods for many sequence-based tasks. Next we describe in detail the Transformer Multi-Head Attention architecture.

A single Transformer attention block has three tensor inputs denoted as query $Q$, key $K$ and value $V$ with dimensionality of $d_q, d_k$, and $d_v$ respectively. The output is a weighted $V$ tensor with weights calculated by the compatibility function of the $Q$ and $K$.

\begin{equation}
\label{eq:attention}
\text{Att}(Q,K,V)=\text{Softmax}(\frac{QK^{T}}{\sqrt{d_k}})V\\
\end{equation}

where dimension $d_k=d_q$ of keys and queries is used as a scaling factor to prevent large values of the dot product.

But, instead of using a single attention block, $Q$, $K$, and $V$ are linearly projected $h$ times with different separate neural layers $W^Q_i$, $W^K_i$, and $VW^V_i$. 
Each set of projections is passed as input to the attention block with $d_v$-dimensional output. 

\begin{equation}
H_i=\text{Att}(QW^Q_i,KW^K_i,VW^V_i),
\end{equation}

All outputs values of each projection then concatenated and  projected by $W^O$ neural network, giving us the output of the multi-head attention block.

\begin{equation}
\text{MultiHead}(Q,K,V)=\text{Concat}(H_1,\dots,H_h)W^O\\
\end{equation}

We incorporate this Multi-Head Attention block architecture to create a Transformer Tracker network in Section~\ref{sec:network}.

\section{DOTCL}

This section introduces our end-to-end Deep Object Tracking with Circular Loss Function approach and describes the network architecture, the new bounding box parametrization, and the new loss function.

\subsection{Transformer Tracker Architecture}\label{sec:network}

\begin{figure}[ht]
\centering     
\includegraphics[width=0.8\linewidth]{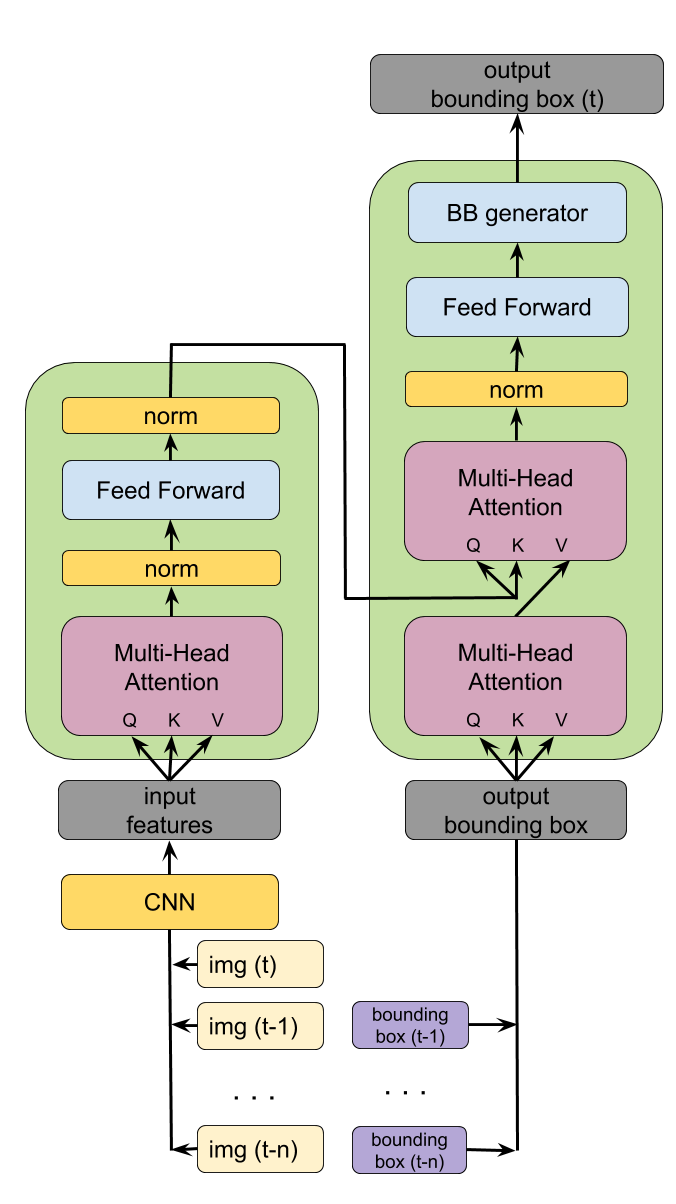}
\caption{Our Transformer Tracker (TT) architecture consists of the encoder (left) and the decoder (right). The encoder receives several previous frames including the current one. The decoder receives bounding boxes from previous steps and the output of the encoder and predicts the bounding box for the current frame. See Section~\ref{sec:network} for more details.}
\label{img:network}
\end{figure}

Figure~\ref{img:network} shows the overall network architecture dubbed Transformer Tracker (TT). The left part is the encoder network and the right part is the decoder network.

The encoder receives $n$ previous frames plus the current one. Convolutional neural network (CNN) extracts high-level features. The CNN is pretrained for image classification task on a large dataset (i.e., ImageNet\cite{russakovsky2015imagenet}). These input features are then passed to multi-head attention block (described in Section~\ref{sec:transformer}) and, subsequently, to fully connected feedforward network.

The output of the encoder is then passed to the second multi-head attention block of the decoder network.

The decoder receives predicted bounding boxes from the previous $n$ steps and passes them to the its first multi-head attention block. The output of the first multi-attention block is passed to the second multi-head attention block along with the output from the encoder network. The output of the second multi-head attention block is passed to the feedforward network and then to the bounding box generator network for the generation of the bounding box parameters.

In the original Transformer, keys $K$ and values $V$ for the second multi-head attention block of the decoder are the output of the encoder, and the queries $Q$ were the output of the first multi-attention block of the decoder itself. In our architecture, we use the output of the first multi-head attention block as values $V$ for the second multi-head attention block, since each next box should depend on the locations of the previous ones.

We also found that performing layer normalization for select layers (Figure~\ref{img:network}) improves performance of the network. 

This architecture allows us to access both the previous $n$ frames as well as their predicted bounding boxes which seems to benefit the overall performance.

When we initialize the network at the first frame of the sequence, we copy this first frame and the ground truth bounding box $n$ times. These copies in a sense serve as previous frames and bounding boxes.

This completes a general description of our Transformer Tracker architecture. We give technical details of implementation in Section~\ref{sec:training}.

\subsection{Parametrization for the Rotated Bounding Box}\label{sec:parametrization}

Previous state-of-the-art methods employ two main approaches for the parameterization of the bounding box:

\begin{itemize}
    \item Eight-dimensional vector with $x$ and $y$ coordinates of the four vertices of a polygon.
    \item Four-dimensional vector with $x$ and $y$ coordinates of the center, height $h$ and width $w$ of a rectangular bounding box. This parametrization only allows for axis-aligned bounding boxes.
\end{itemize}

It is important to note that out of the top three approaches of the VOT2018 competition, only one (LADCF) used the eight-dimensional parametrization. Even though the ground truth was a rotated bounding box, most current state-of-the-art solutions (including the only end-to-end deep learning solution) opt for the non-rotated predicted bounding box. This is due to a more straightforward training and tuning process.

\begin{figure}[ht]
\centering     
\includegraphics[width=0.5\linewidth]{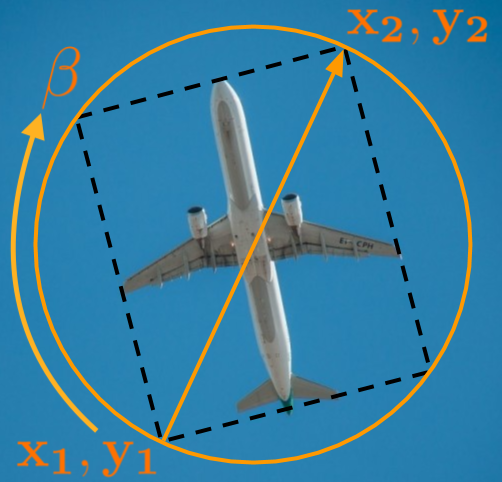}
\caption{New five-dimensional bounding box parametrization (5BB). We parametrize the bounding box by the coordinates of two diagonal points ($x_1$, $y_1$) and ($x_2$, $y_2$) and the normalized length ($\beta$) of the arc from $(x_1,y_1)$ to the next angle of the box in the clockwise direction.}
\label{img:params}
\end{figure}

We introducе a new parametrization of the bounding box that we argue is better suited for the rotated variant (Figure~\ref{img:params}).

Let us denote the coordinates of two diagonal points of the bounding box as ($x_1$, $y_1$) and ($x_2$, $y_2$). Then all rectangular bounding boxes with those two diagonal points will be inscribed in the circle with the $((x_1,y_1),(x_2,y_2))$ diameter. That means we can define the exact bounding box by only one more parameter --- the normalized length of the arc from the $(x_1,y_1)$ point to the next point in the clockwise direction (see Figure~\ref{img:params}). This parameter ($\beta$) also characterizes the proportions of the bounding box.

We then can use this five-dimensional parametrization of the bounding box (5BB) with $\text{Smooth}_{L_1}$ loss function (Equation \ref{eq:smooth}). As an additional improvement, we propose a better-suited loss function.

\subsection{Circular Loss Function}\label{sec:loss}
Our bounding box parametrization naturally induces a loss function that consists of three parts.

\begin{equation}
\label{eq:total_loss}
    L=L_{area}+\lambda_1L_{angle} + \lambda_2L_{arc}
\end{equation}

The first part is the inverted relative overlap of the circumscribed circles of the predicted ($C$) and the ground truth ($C_\text{gt}$) bounding boxes. This circle overlap serves as a crude differentiable approximation of the bounding box overlap.  

\begin{equation}
L_{area} =  1 -  \frac{|C \cap C_{\text{gt}}|}{|C \cup C_{\text{gt}}|}
\end{equation}

We can calculate the overlap of the circles as a function of the total area $S$ and the intersection area $I$.

\begin{equation}
    \label{eq::circle_relative_intersection}
    \frac{|C \cap C_{\text{gt}}|}{|C \cup C_{\text{gt}}|} = \frac{I}{S - I} \\
\end{equation}

We can then, in turn, calculate the intersection area $I$ according to the following equation: 

\begin{equation} \label{eq::circle_intersection}
\begin{split}
    &I = r^2 \arccos{\left( \frac{d^2 + r^2 - r_{\text{gt}}^2}{2 d r} \right)} \\       
    &+ r_{\text{gt}}^2 \arccos{\left( \frac{d^2 + r_{\text{gt}}^2 - r^2}{2 d r_{\text{gt}}} \right)} \\
    \medskip
      &- \frac{1}{2} \sqrt{(d + r_{\text{gt}} + r) (d + r_{\text{gt}} - r) (d + r - r_{\text{gt}}) (r + r_{\text{gt}} - d)}
\end{split}
\end{equation}

where $r$ and $r_{\text{gt}}$ are the radii of the predicted and the ground truth circles respectively, $d$ is the distance between circle centers. These values are easily expressed through $x_1, y_1, x_2, y_2$.

Calculation of the total area of the circles is a trivial task.

\begin{figure}[ht]
\centering
\includegraphics[width=0.5\linewidth]{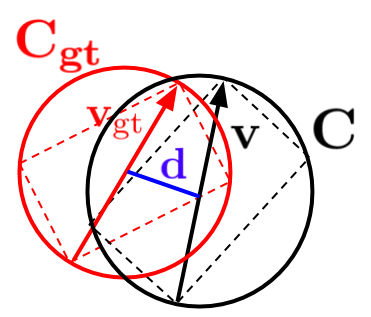}
\caption{We devise the loss function that takes into account the overlap of circumscribed circles and relative orientation of the ground truth and the predicted bounding boxes. See Section~\ref{sec:loss}.}
\label{img:loss}
\end{figure}

The second part reflects the angle between the vector $\mathbf{v}$ (Figure~\ref{img:loss}) of the predicted bounding box and the corresponding vector $\mathbf{v_\text{gt}}$ of the ground truth bounding box. This part of the loss function aims to ascertain the mutual rotation of the predicted and ground truth bounding box.

\begin{equation}
L_{angle} = 1 - \cos{ \angle (\mathbf{v}, \mathbf{v_{\text{gt}})}}
\end{equation}

The last part reflects how well we predicted the proportions of the bounding box. It is formulated as the squared difference between the normalized lengths of the arcs from the $(x_1,y_1)$ point to the next point in the clockwise direction ($\beta$ on Figure~\ref{img:params}).

\begin{equation}
L_{arc} = (\beta - \beta_{\text{gt}})^2
\end{equation}

This circular loss function (CL) is differentiable in terms of $x_1, y_1, x_2, y_2$, and $\beta$. Therefore, we can calculate the gradient and use the backpropagation method to train our network. For more details of our implementation, we refer the reader to source code available at [link is removed for review purposes].

The coefficients in the sum of loss functions in the Equation~\ref{eq:total_loss} can be used to increase the emphasis on one of the geometric aspects of the positioning of the predicted bounding box.

\section{Experimental Setup}

In this section, we provide the technical details of our implementation. First, we describe the datasets that were used to train the network and evaluate the performance. Next, we specify the parameters of our architecture. And lastly, we describe the additional techniques that we used to improve the performance.

\subsection{Datasets}
We evaluate the performance of our tracking method on the benchmark dataset VOT2018. The main difference between this dataset and other object tracking datasets is that it allows the bounding box for the object to be rotated.

VOT2018 dataset consists of 60 videos of 40 to 1500 frames in length. We randomly split them into 30 videos for training and 30 for testing.  Along with 30 training videos from VOT2018 dataset we also trained our network on Davis \cite{Davis2017}, OTB50 and OTB100 datasets \cite{wu2015object}. Even though these datasets only include axis-aligned bounding boxes, we found it to be of use for the additional refinement.

\subsection{Evaluation}
\label{sec:evaluation}

We evaluate the performance in terms of accuracy (A), robustness (R) and expected average overlap (EAO). Accuracy (A) is defined as the average overlap between the predicted and ground truth bounding boxes.

Robustness (R) is defined as:
\begin{equation}
\text{R} = 100 \frac{N_{\text{fails}}}{N_{\text{frames}}}
\end{equation}

where $N_{\text{fails}}$ is the number of frames with zero overlap. In other words --- a complete failure of the tracker. $N_{\text{frames}}$ is the total number of frames.

The VOT challenge evaluation algorithm uses reset based methodology. Whenever a tracker predicts a bounding box with zero overlap, a failure is detected, and the tracker is reinitialized in five frames after the failure. Plus, ten frames that follow the failure do not count towards accuracy. 

The overall performance is evaluated using the expected average overlap (EAO) which takes account both accuracy and robustness. The detailed description of this metric is quite complex in our opinion does not fit in the scope of this paper. For the complete explanation of the EAO metric, we refer the reader to the original paper \cite{kristan2015visual}.

\subsection{Training}\label{sec:training}
This section lists parameters and training regimes that demonstrated the best performance.

\textbf{CNN:} We choose ResNet-50 \cite{he2016deep} pre-trained on ImageNet dataset \cite{russakovsky2015imagenet} as a feature extractor as it proved to be efficient in many computer vision tasks. This CNN takes a video frame (resized to 224x224 resolution) as an input. We add a layer of $d_{model} = 1024$ output neurons that serve as input tensors (Q, K, V) of the encoder network.

\textbf{Transformer Tracker:} At each step, the encoder network takes the last $n=7$ frames and the current frame represented as 1024-dimensional encodings (output of the CNN). The decoder receives the eight- or five-dimensional (5BB) parametrization of the previous $n=7$ predicted bounding boxes. Both encoder and decoder multi-head attention blocks consist of $h=4$ individual attention heads.  As for the dimensions of Q, K, and V, they are induced by the dimensionality of inputs and outputs. We used 1024x8-dimensional tensors for the first multi-head attention block of the encoder, 5x7-dimensional tensors for the first multi-head attention block of the decoder, and 1024x8-dimensional tensors for the second multi-head attention block of the decoder.

\textbf{Model Training:} At the training phase, we fix the CNN weights, while learn the Transformer weights through backpropagation.  Parameter exploration led us to $\lambda_1 = 0.5$ and $\lambda_2 = 0.3$ coefficients for the total loss function (see Equation~\ref{eq:loss_total}). Each batch consisted of 20 consecutive frames. We use PyTorch library to code all our networks and the included Adam optimizer for 150 epochs with exponentially annealed by 0.94 learning rate starting from $10^{-3}$ \cite{kingma2014adam}.  We performed the training on one NVIDIA Tesla P100 GPU. 150 epochs of training take $\approx36$ hours.

\subsection{Additional techniques}

\subsubsection{Augmentation}

We used the following commonly used data augmentation techniques.  

\begin{itemize}
\item Flip: the video is horizontally and vertically flipped.
\item Blur: blur with a Gaussian filter.
\item Rotation: rotation from a fixed set of 12 angles ranging from -60 to +60 degrees.
\item Brightness shift: random shift in brightness.
\item Contrast shift: random shift in contrast.
\item Color shift: random shift in color scheme.
\end{itemize}

We used the PIL python library\cite{pil2016} to apply these augmentations. Most of the state-of-the-art solutions perform the same data augmentation.

These augmentation techniques extended the training dataset to 13500 total videos.

\subsubsection{Crop}

Cropping the frames is often beneficial for model performance \cite{li2018high} since the size of the object is commonly small relative to the size of the frame.

We crop the frame to the area around the bounding box extended by 1.5 times the diagonal length. Again, most of the other state-of-the-art approaches employ cropping in some form.

\subsubsection{Pretraining}
\label{sec:pretraining}
According to the algorithm of the VOT2018 challenge, if the model loses an object, no predictions are accepted for five frames. After five frames, the model receives new ground truth bounding box and starts over. To use the newly provided ground truth bounding box with maximum benefit, we have tuned the re-initialization of the tracker.

To achieve that, we use the first 15 epochs for re-initialization training with just the $\text{Smooth}_{L_1}$ metric. We assume that tracker loses the object and re-initialization occurs every frame. Fixing a specific Transformer Tracker architecture, you can do this pre-training once and then use the weights to initialize the neural network in the main learning phase.

Thus the model's re-initialization can be organically embedded in its description, so that regardless of the inputs during re-initialization, the network immediately gives a close to correct answer. This allows us to get rid of the stage of coarse adjustment of the model and train the model for the target metric.

\section{Results}
\label{sec:res}

Table~\ref{tbl:results} shows the comparison of different combinations of our methods to state-of-the-art approaches in terms of accuracy (A) and robustness (R), as well as expected average overlap (EAO). MFT and LADCF user discriminative filters with both hand-crafted and deep features from CNN, and SiamRPN is the only other end-to-end deep learning approach. \textbf{TT} denotes the architecture with only the Transformer Tracker network with the standard bounding parametrization and loss. \textbf{5BB} refers to the addition of the new parameterization for the rotated bounding box and \textbf{CL} is our new circular loss function. \textbf{PT} denotes the pretraining process.

\begin{table}
\caption{Results of our model}
\centering
\begin{tabular}{ |l|c|c|c| }
\hline
Model & \textbf{A}  & \textbf{R} &  \textbf{EAO}   \\
\hline
MFT & 0.493 & 0.136 & 0.383 \\
LADCF & 0.498 & 0.145 & 0.390 \\
\hline
SiamRPN & 0.569 & 0.272 & 0.379 \\
TT & 0.474 & 0.207 & 0.322 \\ 
TT+5BB & 0.489 & 0.197 & 0.335  \\ 
TT+5BB+CL & 0.515 & 0.168 & 0.373 \\
DOTCL (TT+5BB+CL+PT) & 0.533 & 0.151 & \textbf{0.396} \\
\hline
\end{tabular}
\label{tbl:results}
\end{table}

As we can see from the table, Transformer Tracker (TT) approach alone allows us to improve the robustness (R) of the end-to-end approaches by 25\%. With new bounding box parametrization, circular loss function, and pretraining, our model (DOTCL) outperforms both discriminative filter and end-to-end state-of-the-art approaches in terms of the expected average overlap (EAO) metric. It also demonstrates the improvement of 44\% in robustness when compared to the end-to-end deep learning method (SiamRPN).

\section{Conclusion}

In this paper, we present a novel end-to-end deep learning approach to visual object tracking task. It utilizes Transformer Tracker architecture to interpret spatiotemporal relations. We also introduced a new type of bounding box parametrization and loss function that is aimed to work well in the case of the rotated bounding box. 

This method (DOTCL) outperforms other state-of-the-art approaches on VOT2018 challenge dataset in terms of expected average overlap metric and improves the robustness metric by 44\% relative to the previous best end-to-end deep learning state-of-the-art approach (SiamRPN).

\bibliographystyle{unsrt}
\bibliography{egbib}

\end{document}